\documentclass[10pt, conference, compsocconf]{IEEEtran}
\pdfoutput=1
\renewcommand\IEEEkeywordsname{Keywords}
\usepackage{cite}
\usepackage{amsmath}
\usepackage{cases}
\usepackage{graphicx}
\usepackage{multirow}
\usepackage{subfig}
\begin{document}
	\title{Multivariate Forecasting of Crude Oil Spot Prices using Neural Networks}
	\author{\IEEEauthorblockN{Ganapathy S. Natarajan}
		\IEEEauthorblockA{School of Mechanical, Industrial, and Manufacturing Engineering\\
			Oregon State University\\
			Corvallis, OR, USA\\
			Email: gana.natarajan@oregonstate.edu}
		\and
		\IEEEauthorblockN{Aishwarya Ashok}
		\IEEEauthorblockA{Department of Computer Science and Engineering\\
			University of Texas at Arlington\\
			Arlington, TX, USA\\
			Email: aishwarya.ashok@mavs.uta.edu}
	}
\maketitle
\begin{abstract}
	Crude oil is a major component in most advanced economies of the world. Accurately predicting and understanding the behavior of crude oil prices is important for economists, analysts, forecasters, and traders, to name a few. The price of crude oil has declined in the past decade and is seeing a phase of stability; but will this stability last? This work is an empirical study on how multivariate analysis may be employed to predict crude oil spot prices using neural networks. The concept of using neural networks showed promising potential. A very simple neural network model was able to perform on par with ARIMA models - the state-of-the-art model in time-series forecasting. Advanced neural network models using larger datasets may be used in the future to extend this proof-of-concept to a full scale framework.
\end{abstract}
	\section{Introduction}
		Crude oil spot prices saw a tremendous up-tick in the first decade of the 21\textsuperscript{st} century. Since 2014, crude oil prices have fallen and may have stabilized now. However, there has always been a constant interest in accurately predicting crude oil prices; given that crude oil drives a major portion of the economy. Economists, scientists, data analysts, and traders are all interested in models that give them the best accuracy. In the last decade, advances in machine learning has enabled everyday data analysts to use techniques like neural networks; case-in-point is Google's TensorFlow\textsuperscript{TM}. 
		
		Crude oil prices have mostly been forecasted using time-series methods. There is some work in dealing with crude oil price forecasting as an econometric problem; however, there is very limited work in using multivariate techniques that move away from traditional regression modeling. This paper combines the lack of multivariate forecasting and advances in machine learning to provide a proof-of-concept for using Neural Networks in multivariate forecasting of crude oil prices. The results presented in this study are mostly empirical and lays the foundation for more in-depth studies in this direction.
	\section{Related Work}
		We submitted a literature review paper to a conference \cite{self:IISE}. This work looked at the different types of models and methods used in forecasting oil prices in general. Time series models were predominantly used. Time series models use just the oil price over time to predict future prices. ARIMA group of methods were the most commonly used time series forecasting methods \cite{moshiri2006forecasting,mostafaei2011modeling,komijani2014hybrid,zou2015wavelet} either for model building or as a benchmark or for a part of the model building. Aritficial intelligence and machine learning based models \cite{bildirici2013forecasting,shu2014multiscale,deng2014crude,yu2016novel,yu2015decomposition} including neural network models \cite{cuaresma2009modelling} were also used to just predict the time series or components of the time series.
		
		The second type of approach to forecasting oil prices was econometric models. Econometric models usually perform linear regression based analysis and take into account economic factors. Some of the common independent variables used were income growth, price increase, and population growth \cite{dahl1994oil}, gas tax exemption \cite{decker2005impact}, OECD stocks, and OPEC spare capacity \cite{merino2010econometric}. OECD is the Organization for Economic Co-operation and Development - a group of developed economies in the world, and OPEC is the Organization of the Petroleum Exporting Countries. Some of the variables used in econometric studies historically, will be used in this study too, viz. CPI, population, petroleum stocks, to name a few.
		
		The third commonly used methods are multivariate models, which are of interest in this paper. Multivariate models use multiple variables akin to econometric models, but do not always assume a linear relationship or use just regression. Some of the commonly used multivariate methods include multivariate time series models \cite{myklebust2010forecasting}, neural networks \cite{movagharnejad2011forecasting,pang2011forecasting}, and multi-criteria decision models \cite{xu2012performance}. Reference \cite{movagharnejad2011forecasting} used neural networks to predict commercial oil price using country of origin, sulfur content, and density, and \cite{pang2011forecasting} used wavelet neural network to predict monthly crude oil price using just the inventory levels - not truly a multivariate methodology.
		
		Recent works have continued with the trend of using different methods but still mostly perform time-series forecasting. Although the recent works use hybrid models \cite{safari2018oil}, ensemble neural network models \cite{ding2018novel}, gray wave model\cite{chen2018multi}, and wavelet neural network models\cite{huang2018global}, they still are time-series forecasting models. Reference \cite{wang2018novel} used network science and predicted parts of a time series to deal with specific non-linear patterns.
		
		The common pattern in all the related work is that most of them used time-series or hybrid models to forecast the time-series; or use econometric or multivariate models using very few variables. This paper aims at exploiting the gap in using multiple variables and move away from econometric models. Recent developments in the accessibility and usability of neural network pushed us to strongly believe that neural network models built using multiple input variables would be the way forward in forecasting crude oil spot prices.
	
	\section{Theoretical Background}
		\begin{figure}[!b]
			\centering
			\includegraphics[width=3in]{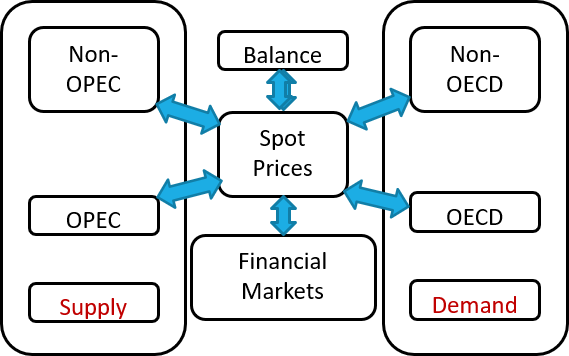}
			\caption{Theoretical Model adapted from \cite{TheoModel:online}}
			\label{fig:theo_full}
		\end{figure}	
		\begin{figure}[!b]
			\centering
			\includegraphics[width=3in]{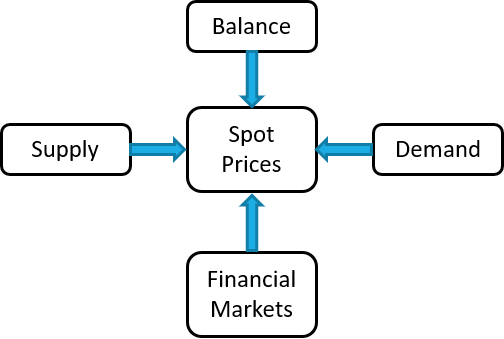}
			\caption{Updated Theoretical Model used in this Study}
			\label{fig:theo}
		\end{figure}	
		The theoretical model used in this research study is based on the Energy Information Administration’s (EIA) model on what drives crude oil spot prices \cite{TheoModel:online}. Fig. \ref{fig:theo_full} shows this model. For the purposes of our exploratory study, the model shown in Fig. \ref{fig:theo_full} is updated to the model in Fig. \ref{fig:theo}. The updated model simplifies the OECD and Non-OECD, and OPEC and Non-OPEC components to just demand, and supply components, respectively. Since our study deals with only the US market and WTI crude oil prices, there is no need for the differentiation. In addition, the original model assumes that spot prices also have an effect on other variables. In an exploratory study, such a relationship would complicate the model. In order to simplify the relationship and make spot price a truly dependent variable, the relationship is modified such that the four categories of variables affect spot prices. These are simplifying assumptions but are necessary for this research. A discussion on how these assumptions may be changed in building models is discussed as future work in the Conclusions section. In this exploratory proof-of-concept research three variables were chosen for each factor (category of variable). The spot price used was the West Texas Intermediate (WTI). A detailed description of the dataset and variables is provided in Section V.
	
	\section{Methods}
		Three methods were used to analyze the data and predict the WTI crude oil prices. Neural network, the concentration of this paper, regression with L2 regularization, and ARIMA methods were used. Regression and ARIMA were performed to compare and benchmark results. Each of the methods used is explained in the following sections. A section on the different evaluation metrics used will follow these sections.
	
		\subsection{Neural Network}
			
			Neural Networks usually consist of one input layer, one output layer, and a few hidden layers in between the input and the output layers. Each layer consists of weights and biases that are adjusted during the training process. Learning occurs through forward pass over the network and backpropagation of errors based on the forward pass. Some of the important choices and parameters that may be changed include number of hidden layers, number of neurons per hidden layer, choice of loss function, activation function in the different layers, learning rate, and number of iterations. In this study, the number of hidden layers is fixed at two. The number of neurons per hidden layer is twelve – the same as the number of independent variables. The loss function used to calculate the error to be used in backpropagation is Mean Squared Error (MSE). MSE is a widely used error function for continuous numerical data like the WTI.
								
			The hidden layer activation function is chosen to be the Exponential Linear Unit (ELU). ELU activation function is defined as follows:			
			\begin{equation}
			f(x)=\begin{cases}
			\alpha(\exp(x))-1&\mbox{if }x\le0\\
			x &\mbox{if }x > 0\end{cases}
			\label{eq:elu}
			\end{equation}
			\begin{figure}[!t]
				\centering
				\includegraphics[width=2in,height=1.5in]{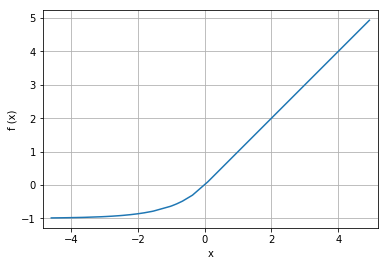}
				\caption{The Exponential Linear Unit Activation Function}
				\label{fig:ELU}
			\end{figure}
			\begin{figure*}[!t]
				\centering
				\includegraphics[width=6in]{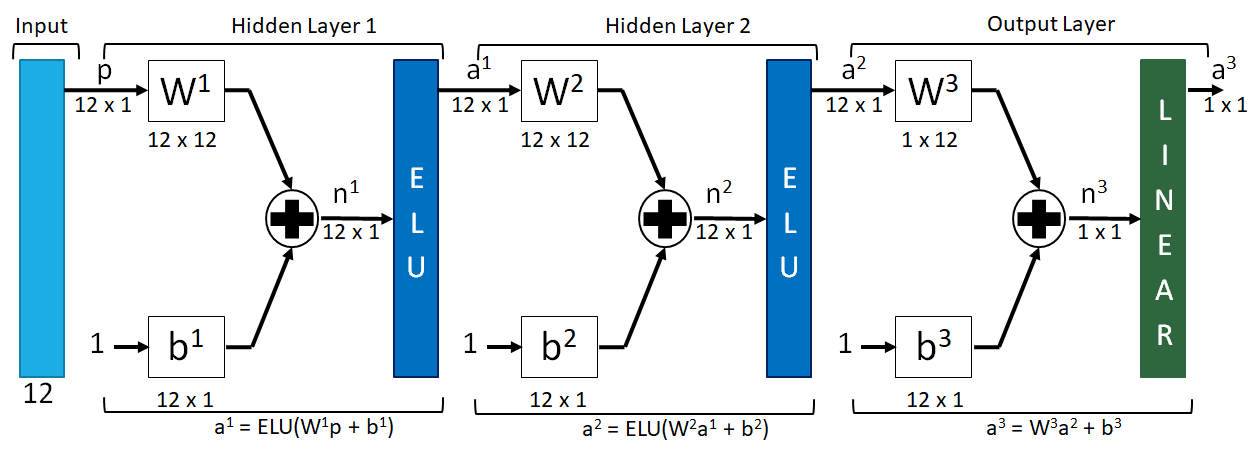}
				\caption{Network Architecture for the Neural Network Method}
				\label{fig:NNet}
				\small \textit{Note: The superscripts refer to the layer and is not a power.}
			\end{figure*}
			
			In this study the $\alpha$ value in (\ref{eq:elu}) is set to 1; effectively, the function for negative values is $(\exp(x) – 1)$. The graph of the activation function is shown in Fig. \ref{fig:ELU}. ELU has been shown as a good activation function for continuous data. The function has a mean centered around zero due to reduction of bias and exponential function around zero and small negative values. The combination of these two factors helps in faster learning when using ELU activation function \cite{DBLP:journals/corr/abs-1804-02763}. The output layer activation function is a linear activation function such that the output may be compared against the actual WTI to calculate the loss function. The complete architecture along with matrix dimensions may be found in Fig. \ref{fig:NNet}.
			
			The last two parameters that can be changed are learning rate and number of iterations. These two variables were experimented on in this study. Large learning rates have the potential of missing the minimum and smaller learning rates have the potential of being stuck in a local minimum. Learning rates of 0.001 and 0.0001 were chosen in this study. In choosing the number of iterations, a concept similar to learning rate applies. Too many iterations may cause the model to be overfitted and too little iterations might mean the learning is not complete. An appropriate number of iterations are needed for the learning to be complete. In this study iterations of 100, 150, and 200 were tried. An iteration is complete when every single data point has been seen by the model. In certain areas this is defined as an epoch. So, 100 iterations would mean the entire dataset was run through the neural network model 100 times.
			\subsection{Regression}
			Multiple linear regression is a common method used when multiple independent variables are involved. Moreover, the regression model is based on reducing the least squares or minimizing mean squared error – the same loss function used in the neural network model here. So, regression model will be used as a means to compare the performance of neural network model.
			
			Regression models tend to overfit as the number of independent variables increase. One method to avoid overfitting is L2 regularization. L2 regularization adds a correction term that is the square of the regression coefficients to the objective function. L2 regularization, referred to as ridge regression, will also help in case of highly correlated independent variables. With L2 regularization, the regression problem is represented by (\ref{eq:reg}).
			\begin{equation}
			min\sum_{i=1}^{n}(y_i-\beta^{T}X_i)^2 + \lambda \sum_{j=1}^{k}\beta_{j}^{2}
			\label{eq:reg}
			\end{equation}
			where,\\
			n – number of data points\\
			k – number of independent variables\\
			$\beta$ – regression coefficients\\
			X – matrix of independent variables\\
			y – matrix of dependent variable\\
			$\lambda$ – regularization coefficient\\
			
			In this study the closed form solution for the ridge regression was implemented. The regression coefficients may be predicted using \ref{eq:regsol}.
			\begin{equation}
			\hat{\beta}=(X^TX+\lambda I)^{-1}X^Ty
			\label{eq:regsol}
			\end{equation}
			where,
			I – identity matrix of size k. The remaining symbols are as identified in (\ref{eq:reg})
			
			The regression model was implemented assuming no constant term. From (\ref{eq:reg}) and (\ref{eq:regsol}), it can be observed that the one parameter that could be changed is $\lambda$. In this study $\lambda$ values of 0 (no regularization), 0.25, 0.5, 0.75, 0.95, and 0.99 were used. Each time the $\lambda$ values were changed a new regression model was generated.
		\subsection{ARIMA}
		\begin{figure}[!t]
			\centering
			\includegraphics[width=3in]{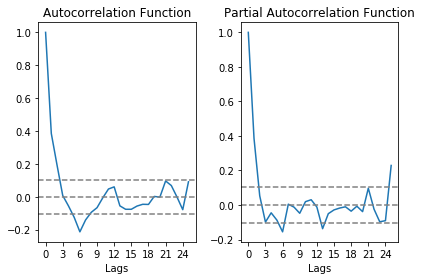}
			\caption{AC and PAC for the time series 1-differenced}
			\label{fig:Diff1}				
		\end{figure} 
		\begin{figure}[!t]
			\centering
			\includegraphics[width=3in]{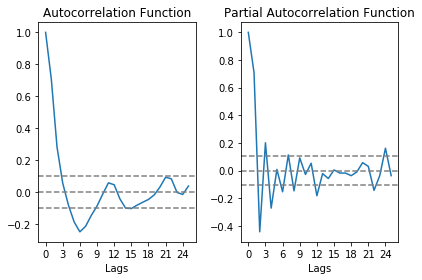}
			\caption{AC and PAC for the time series 2-differenced}
			\label{fig:Diff2}				
		\end{figure}
			ARIMA stands for Auto-Regressive Integrated Moving Average. ARIMA model incorporates a regression component, an integrative component - achieved by differencing the time series, and a moving average component. The number of these components are usually presented as (p,d,q), respectively. There are variants of the ARIMA that does just ARMA or seasonal ARIMA - where there is another set of (p,d,q) to take care of seasonality. In this paper, the ARIMA model will be used as a benchmark. ARIMA does not use the independent variables like the other methods; it uses the times series as the input to build a model. Choosing (p,d,q) is based on a set of rules and the auto-correlation (AC) and partial auto-correlation (PAC) plots. Detailed explanation of the different rules to choose p, d, and q may be found in \cite{ARIMARules:online}.
			
			Usual practice to choose the number of differencing terms is based on if the trend is a constant linear trend or if the trend changes over time. When the trend changes over time, we usually choose d = 2. This means differencing is done by shifting the time series two time periods. For this study, we will try both d = 1 and d = 2. Fig. \ref{fig:Diff1} and Fig. \ref{fig:Diff2} show the AC and PAC plots for d = 1 and d = 2, respectively.
			
			The number of p and q terms are chosen by looking at the PAC and AC plots, respectively. The number of p terms is based on the number of lags where the PAC plot cuts off, or falls below the confidence limit (marked in the figure with a horizontal dotted line). The number of q terms is chosen similarly, but looking at the AC plot. In this study we will experiment with different values of p, d, and q, depending on the AC and PAC plots. Care should be taken to not choose combinations of p and q where the AR term cancels the MA term. So the combinations of (p,d,q) chosen for this study are (1,1,2), (2,1,1), (2,1,3),(1,2,2),(2,2,3), and (2,2,5).

		\subsection{Evaluation Metrics}
			The common error measures used in forecasting are mean squared error (MSE) and mean absolute deviation (MAD). Although MSE is a good measure, it does not put the error in the same units as the forecasted variable. For this reason, we used root mean squared error (RMSE) which is the square root of the MSE. In addition to RMSE and MAD, since we use independent variables to explain the variance in a dependent variable, R-squared value was used as an evaluation metric. R-squared value increases with increase in the number of the independent variables. Therefore, the adjusted R-squared value was used for the in-sample data. Since the weights, regression coefficients, and other parameters are not changed from the in-sample to the out-of-sample data, adjusted R-squared is not reported for the out-of-sample data. A simple R-squared was calculated for the out-of-sample data.
			
			An evaluation metric calculated based on the in-sample (train) and out-of-sample (test) R-squared data called Generalization was also used. Generalization is the ratio of out-of-sample R-squared to in-sample R-squared value. This is a measurement of how well the model building performed. If the model did not overfit and provided a good generalization of the underlying data, generalization should be high. If the model over-learned the in-sample data, it might not perform well on the out-of-sample data. This will result in a low generalization. We are looking at models that have high generalization so that the model can fit/predict future values with lower error.
	\section{Description of Dataset}
		The dataset consists of twelve independent variables and one dependent variable. The twelve independent variables are divided into three variables each for supply, demand, balances, and financial markets. For the purposes of this research study the variables are assumed to be truly independent. However, some of the variables may be related. This relation will be looked at in the Data Characteristics section. The dependent variable is West Texas Intermediate (WTI).
		\begin{table*}[!t]
			\caption{Description of Variables and their Abbreviations}
			\label{tab:Desc}
			\centering
			\includegraphics[width=6in]{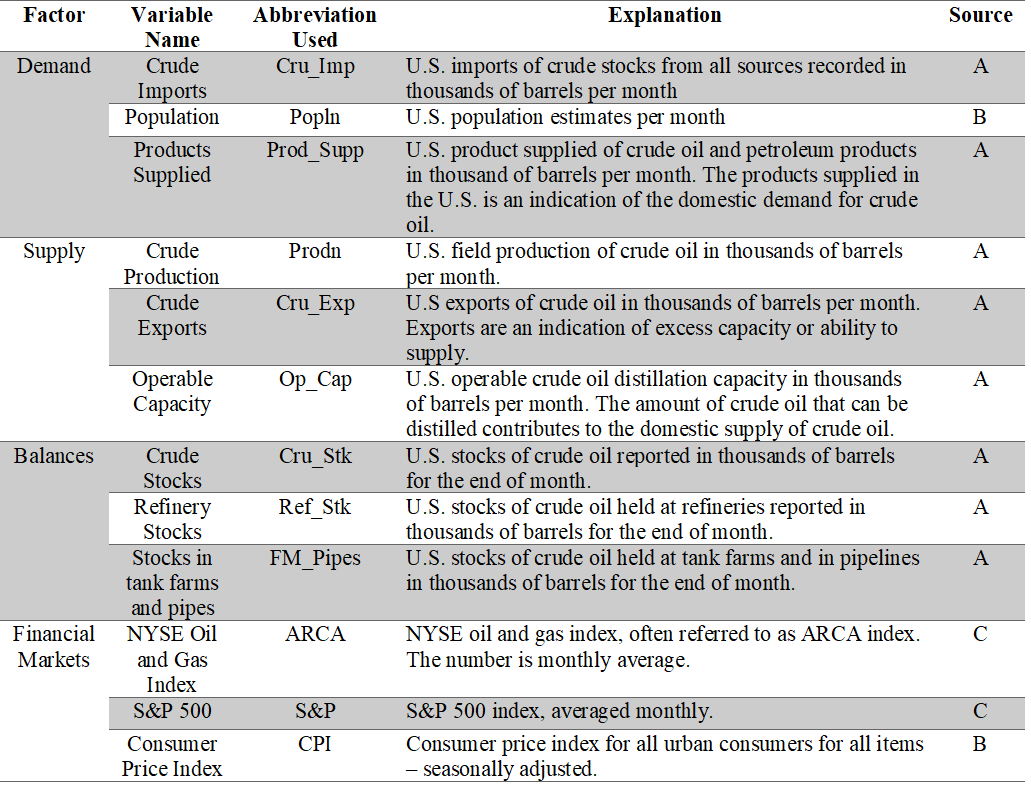}\\
			\raggedright
			\small A - Data obtained from U.S. Energy Information Administration \cite{DataEIA:online}.\\
			\small B - Data obtained from Federal Reserve Bank of St. Louis \cite{Fed}.\\
			\small C - Data obtained from Yahoo! Finance \cite{Yahoo}.
		\end{table*}
		\begin{figure}[!t]
			\centering
			\includegraphics[width=3in,height=2.5in]{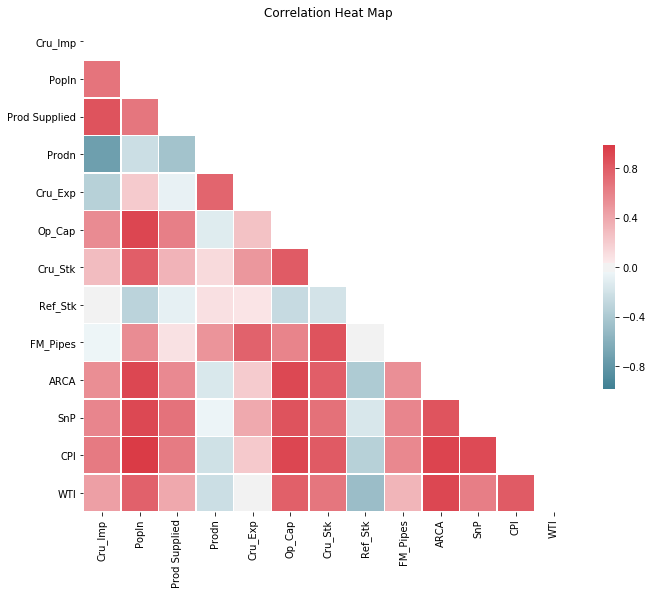}
			\caption{Correlation Heat Map}
			\label{fig:heat}
		\end{figure}
		\subsection{Variables and Definitions}
			Data were collected on a monthly frequency from January 1986 to April 2018. Datapoints from January 2017 to April 2018 were kept aside as out-of-sample (or test) data; the rest of the data is considered in-sample (or train) data. Table \ref{tab:Desc} shows the variables, the category for the variable based on the theoretical model, abbreviation used, short description of variable, and source of data. The data units and scales are vastly different. So, when using for in-sample and out-of-sample tasks, all independent variables were normalized (min-max normalization) to be between 0 and 1 by using (4)
			\begin{equation}
			Normalized Data = \frac{(Data - min.val)}{(max.val - min.val)}
			\end{equation}
			\begin{figure*}[h]
				\centering
				\includegraphics[width=5in,height=4in]{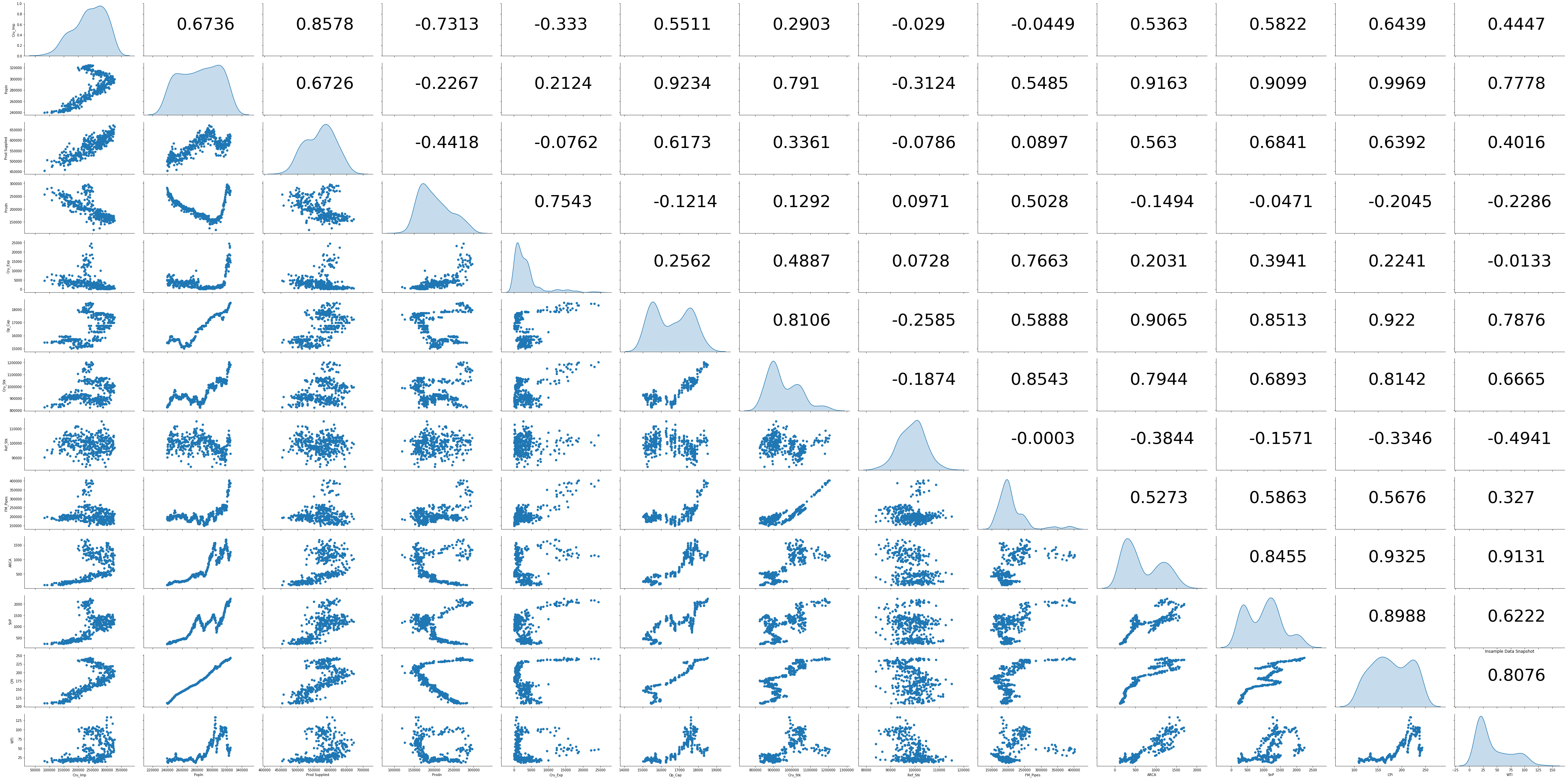}
				\caption{Plot of all Variables with Relationship among Variables, Distributions, and Correlations}
				\label{fig:pair}
			\end{figure*}
		\subsection{Data Characteristics}
			Data characteristics are discussed only for the in-sample (train) data. Fig. \ref{fig:heat} shows the correlation heat map of the independent and the dependent variables used in this study. The variables that have a strong correlation with WTI are ARCA, CPI, Population, and Operable Capacity. Some independent variables show high correlations among themselves. This might cause problems with the regression model. However, with the ridge regression some of these effects may be alleviated. Since this study is exploratory in nature we will not perform feature selection to select the best independent variables.
			
			Fig. \ref{fig:pair} shows the distribution of each of the variables, the scatter plots and correlations. The important thing to notice is how most variables are multi-modal in their distribution. It can be noted that the relationship among the variables is not linear. For these reasons, we hypothesize that linear regression will not perform well with this dataset. This also reaffirms our choice of activation function - an exponential linear unit for the neural network model.
	\section{Results and Discussions}
	\begin{figure*}[!t]
		\centering
		\subfloat[In-sample for Lowest In-sample Error]{
			\includegraphics[width=2.5in,height=1.5in]{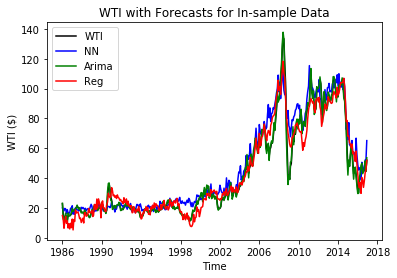}
			\label{fig:trerr_tr}}
		\subfloat[Out-of-sample for Lowest In-sample Error]{
			\includegraphics[width=2.5in,height=1.5in]{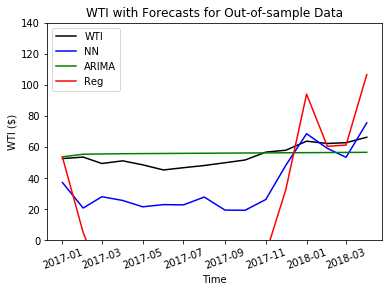}
			\label{fig:trerr_te}}
		\newline
		\subfloat[In-sample for Lowest Out-of-sample Error]{
			\includegraphics[width=2.5in,height=1.5in]{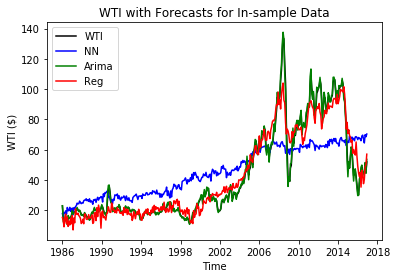}
			\label{fig:teerr_tr}}
		\subfloat[Out-of-sample for Lowest Out-of-sample Error]{
			\includegraphics[width=2.5in,height=1.5in]{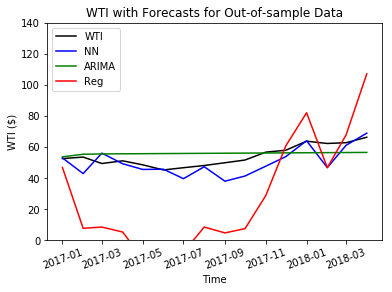}
			\label{fig:teerr_te}}
		\newline
		\subfloat[In-sample for Highest Out-of-sample R-squared]{
			\includegraphics[width=2.5in,height=1.5in]{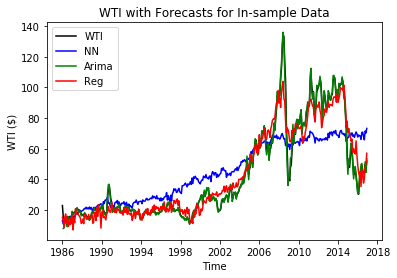}
			\label{fig:ter_tr}}
		\subfloat[Out-of-sample for Highest Out-of-sample R-squared]{
			\includegraphics[width=2.5in,height=1.5in]{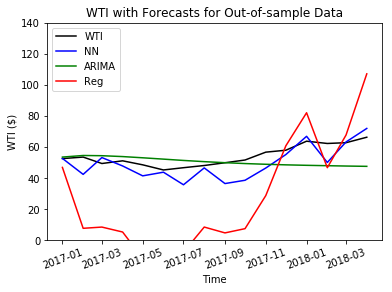}
			\label{fig:ter_te}}
		\newline
		\subfloat[Generalization by Method]{
			\includegraphics[width=2in,height=1.5in]{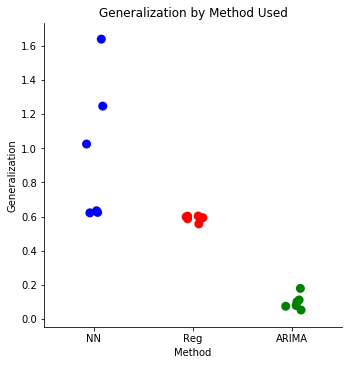}
			\label{fig:Gen}}
		\subfloat[Time by Method]{
			\includegraphics[width=2in,height=1.5in]{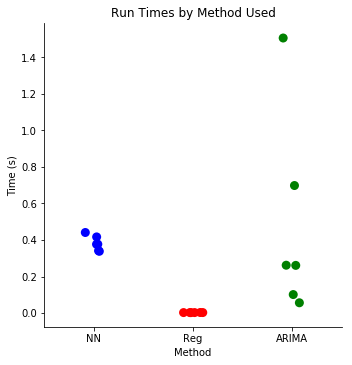}
			\label{fig:Time}}
		\caption{Results with Respect to Different Evaluation Metrics and Run Time}
	\end{figure*}
		Results of the analysis will be discussed based on the evaluation metrics explained in section IV.
		\subsection{Performance based on Error Measures}
			The error measures used were RMSE and MAD. In all the different experiments RMSE and MAD performance matched, i.e. if RMSE was lowest for a certain combination of run, then MAD was also lowest for that same run. Fig. \ref{fig:trerr_tr} and Fig. \ref{fig:trerr_te} show the in-sample and out-of-sample run plots with the best models in terms of the lowest in-sample errors. The lowest in-sample errors were for the following parameters: NN – learning rate of 0.001 with 200 iterations, ARIMA – (2,1,3), and Regression – lambda = 0. ARIMA and Neural Network (NN) performs better than Regression models. ARIMA model has the best performance among the three models. However, looking at the out-of-sample plot, it can be observed that ARIMA just predicts a slowly increasing trend line over the out-of-sample data. This is because ARIMA just uses the time-series and needs to use forecasted values for forecasts further in the future. The NN model on the other hand performs better with following the rise and fall in WTI prices, albeit with some error. Regression is the worst performing as it even predicts some WTI values to be negative, which is not possible. The figures have the same scale on the y-axes for ease of comparison.
			
			Fig. \ref{fig:teerr_tr} and Fig. \ref{fig:teerr_te} show the in-sample and out-of-sample run plots with the best models in terms of the lowest out-of-sample errors. The lowest out-of-sample errors were for the following parameters: NN – learning rate of 0.0001 with 100 iterations, ARIMA – (1,1,2), and Regression – $\lambda$ = 0.99. The NN model on the in-sample looks like a bad fit to the in-sample data. However, the out-of-sample data’s fit is better. The model seems to more closely follow the change in WTI prices. ARIMA model behaves the same way as before: over-fitting in the in-sample data and not able to model the change in WTI prices in the out-of-sample data.
		\subsection{Performance based on R-squared}
			The in-sample error performance mirrors the in-sample R-squared values. Wherever the in-sample error was lowest, the in-sample R-squared values were the highest. So, we will look at the out-of-sample performance based on R-squared values. Fig. \ref{fig:ter_tr} and Fig. \ref{fig:ter_te} show the in-sample and out-of-sample run plots with the best models in terms of the highest out-of-sample R-squared values. The highest out-of-sample R-squared values were for the following parameters: NN – learning rate of 0.0001 with 150 iterations, ARIMA – (2,2,5), and Regression – $\lambda$ = 0.99. Similar to the error performance, the NN model seem to be a bad fit with the in-sample data. However, the out-of-sample forecast follows the rise and fall of the data.
		\subsection{Diebold-Mariano Test}
			We may look at the raw error numbers and the plots for how good or bad the methods performed, but they are statistically insufficient. The Diebold-Mariano (DM) Test \cite{diebold2002comparing} compares two forecasts for a given time series to statistically determine if the two forecasts are different or not. The DM test is usually performed using the modification suggested by Harvey \cite{harvey1997testing}. Mean Squared Error was used to define the loss function in the DM test, to be consistent with the error functions used in building the models. A h value of 1 was used indicating that the forecasts were just 1 step ahead forecasts. The null hypothesis in the DM test is that the two forecasts are not different.
			
			The p-value between the best out-of-sample forecasts of the neural network and regression models was 0.0001. Since the p-value is $<$ 0.05, the null hypothesis may be rejected. The forecasts provided by the neural network model is different from the regression model and based on the error value, the neural network model performs better, statistically, at $\alpha = 0.05$. The p-value for the best out-of-sample forecasts of the neural network and ARIMA was 0.6469. The null hypothesis cannot be rejected and we can conclude that the neural network forecasts are on par with the forecasts from the ARIMA model, statistically.

		\subsection{Generalization}
			Generalization was calculated as the ratio of out-of-sample R-squared and in-sample R-squared. Fig \ref{fig:Gen} shows the generalization for the different methods. In terms of generalization, i.e. applying the learned model to previously unseen data, neural network models outperformed ARIMA models. The variance in the generalization is high in the neural network models, but on average the generalization was superior to the other two methods. In other words, neural network models were able to adapt to the change in trend of the out-of-sample data better than the other two models. The reason for the better performance than ARIMA is that neural network uses information from the multiple variables, whereas, ARIMA has to rely on recently forecasted value for the next one-step ahead forecast. This is one of the major reasons multivariate forecasting must be considered when the dependent variable may depend on multiple factors.
		\subsection{Time Trade-off}
			Neural Network models perform on par with ARIMA models in terms of error and performs better than ARIMA models in terms of generalization. Regression models always perform poorly for the given dataset. Then do neural network models take too much time to run due to the number of iterations and independent variables? Fig. \ref{fig:Time} shows the run times in seconds for each of the three methods. Time was calculated as an average of 25 runs of each model for each combination of parameters experimented with. The time reported is the total time to learn the in-sample data and forecast the out-of-sample data. All experiments were run on a Intel\textsuperscript{R} Core\textsuperscript{TM} i5-7200 CPU @ 2.50 GHz processor with a 8.00 GB installed RAM.
			
			Regression ran in the shortest time, given there are no iterations and we are using the closed-form solution. The calculation only entailed matrix multiplication and inversion. Of the more complex multiple iterations models, neural networks performed better than ARIMA two out of the six runs and worse the other four times. However, the run times for the NN models were mostly consistent and did not vary hugely based on the number of iterations. The time performance for neural network models may be improved by early stopping based on the best model. ARIMA was modeled using an existing library that performs early stopping based on the maximum likelihood estimation. The time vs. generalization trade-off was better for the neural network models.
	\section{Conclusions and Future Work}
		Neural network based modeling showed better accuracy than regression and on par accuracy as ARIMA models. In terms of generalization and ability to follow changing trends in the out-of-sample forecast, neural network was a clear winner. Neural network shows good potential to be used in multivariate forecasting of crude oil prices.
		
		The NN models used in this research were simple feed forward models. Advanced models like recurrent neural networks (RNN) and long short term memory (LSTM) neural network models may be employed in the future. RNN and LSTM may be more suitable for time series forecasting due to their ability to recurrently look at past data points when learning new data points.
		
		The theoretical model in this research assumed no interactions among the independent variables and that the dependent variable did not affect the independent variables. In reality, some of the financial market variables are dependent on crude oil spot prices. So, these variables may actually have a lag with the time series data. These lags must be adjusted and data must be preprocessed to address some of these issues.
		
		Work will be done in the future to add variables regarding news and current events. For example the oil embargo of 1973 had adverse effects on crude oil spot prices. Other than major events like the embargo, smaller events like oil spills, OPEC capping supply, affects crude oil spot prices. These adverse events will be coded into the data based on text based analysis of social media feed and news corpus to incorporate another level of detail into the independent variables.
		
		There were just three variables selected in each category to prove that the concept of multivariate forecasting using neural networks would empirically work. Work is on-going in increasing the number of variables and performing feature selection. Using the best features would improve the accuracy of the models.
	\bibliographystyle{IEEEtran}
	\bibliography{IEEEabrv,ms}			
			
\end{document}